%% file: main.tex
\documentclass{article}

\input{metadata}
\usepackage{paperstyle}
\usepackage{macros}
\crefname{assumption}{assumption}{assumptions}
\Crefname{assumption}{Assumption}{Assumptions}
\begin{document}

\twocolumn[
  \icmltitle{\papertitle}
  \icmlsetsymbol{equal}{*}
  \begin{icmlauthorlist}
    \paperauthors
  \end{icmlauthorlist}
  \paperaffiliations
  \papercorrespondence
  \icmlkeywords{\paperkeywords}
  \vskip 0.3in
]

\paperaffiliationnotice

\begin{abstract}
  \input{sections/abstract}
\end{abstract}

\input{sections/introduction}
\input{sections/setup}
\input{sections/theory}
\input{sections/method}
\input{sections/experiments}
\input{sections/related-work}
\input{sections/conclusion}

\section*{Impact Statement}
\input{sections/impact}

\ifpaperreview\else
  \input{sections/acknowledgements}
\fi

\bibliographystyle{icml2024}
\bibliography{references}

\clearpage
\appendix
\onecolumn
\input{appendix}

\end{document}

%% file: metadata.tex
\providecommand{\papermode}{preprint}

\newcommand{\papertitle}{Predict the Reward, Not the World:\texorpdfstring{\\}{ }World Reward Models for ML Agents}
\newcommand{\paperrunningtitle}{Predict the Reward, Not the World: World Reward Models for ML Agents}
\newcommand{\papersubject}{World reward models for execution-free action selection in LLM agents}
\newcommand{\paperkeywords}{world models, reward models, LLM agents, machine learning engineering, calibration, selective prediction}
\newcommand{\papernotice}{Preprint.}

\newcommand{\paperauthors}{%
  \icmlauthor{Yuxiao Lee}{equal,hkustgz}
  \icmlauthor{Zherui Yang}{equal,hkustgz}
  \icmlauthor{Fan Liu}{hkustgz}
  \icmlauthor{Hao Liu}{hkustgz}
}
\newcommand{\paperaffiliations}{%
  \icmlaffiliation{hkustgz}{The Hong Kong University of Science and Technology (Guangzhou), Guangzhou, China}
}
\newcommand{\papercorrespondence}{%
  \icmlcorrespondingauthor{Hao Liu}{liuh@ust.hk}
}
\newcommand{\paperauthornote}{\icmlEqualContribution}

\newcommand{\paperprojecturl}{}
\newcommand{\paperprojectlabel}{}
\newcommand{\papercodeurl}{}
\newcommand{\papercodelabel}{}

%% file: sections/abstract.tex
Search-based machine-learning (ML) agents improve a solution by writing
candidate programs, executing them, and keeping the best. Execution gives the
search reliable feedback, but a single run can take hours, and most runs end
in programs the agent discards. Recent agents try to save these runs by
letting a large language model act as a world model that predicts how each
program would perform. However, a wrong prediction feeds the search false information. We
argue that an agent does not need to predict the outcome of execution, only
the decision that execution would support. We propose World Reward Models
(WRMs), which learn from real execution results to choose directly which
candidate will perform better, without predicting any intermediate outcome.
We prove that a selector which predicts execution from the same inputs cannot
outperform the best direct selector, that predicted outcomes add an error
which can make a likelihood-trained world model prefer the worse program on
every instance, and that cross-entropy on executed winners drives a WRM toward
the optimum. Our WRM, ForeCritic, adapts a 9B typed decision model to 5k
executed program pairs. On 12,438 held-out pairs from 26 ML tasks, it picks
the better program 79.4\% of the time, while general-purpose LLM judges reach
53--62\%, and its probabilities are calibrated. Inside AIDE, ML-Master, and
ForeAgent, it cuts mean task time by 4.4--7.5$\times$ while keeping their mean
scores within 0.01 or raising them.

%% file: sections/introduction.tex
\section{Introduction}
\label{sec:introduction}

\begin{figure}[t]
  \centering
  \includegraphics[width=\linewidth]{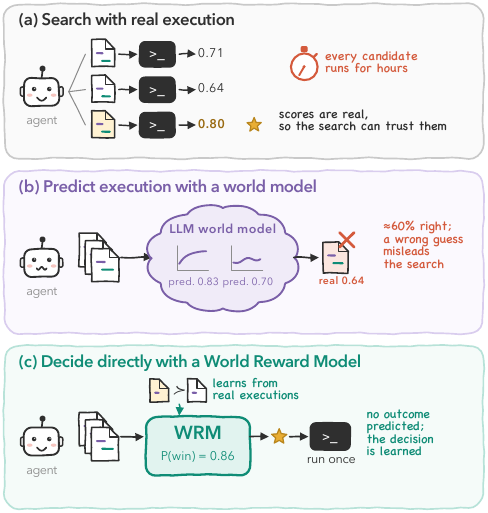}
  \caption{\textbf{Three ways to choose which program to keep.}
    (a)~Executing every candidate gives real scores but costs hours per
    program. (b)~A world model predicts execution instead; its predictions
    are right on about 60\% of program pairs, and a wrong one misleads the
    search. (c)~A World Reward Model learns from real executions to decide
    directly which program will score higher, and the agent runs only the
    program it keeps.}
  \label{fig:teaser}
\end{figure}

Machine-learning (ML) engineering agents treat a task as a search over
programs. AIDE~\citep{jiang2025aide} and ML-Master~\citep{liu2025mlmaster},
for example, draft candidate solutions, execute each one to measure its
validation score, and use the scores to decide what to refine next. Real
execution makes this search reliable, because the scores come from the data
rather than from the agent's own opinion. It also makes the search slow. A
single run can take hours on MLE-bench tasks~\citep{chan2025mlebench}, and
most runs end in programs the agent discards (\Cref{fig:teaser}a).

A natural way to save these runs is to predict them. Model-based
reinforcement learning plans with learned models of the
environment~\citep{sutton1991dyna,schrittwieser2020muzero,hafner2023dreamerv3},
and recent LLM agents use a language model as a world model that anticipates
the result of an action~\citep{hao2023rap,gu2024webdreamer,chae2025wma}.
\foreagent~\citep{zheng2026foreagent} brings this idea to ML engineering. An
LLM reads the task, a data report, and two programs, reasons about how each
would train, and predicts the winner. Only that program is then executed
(\Cref{fig:teaser}b).

These predictions are weak, and weak predictions do harm. With reasoning
enabled, DeepSeek-V3.2 and GPT-5.1 pick the better of two programs 61.5\% and
58.8\% of the time, and accuracy stops improving beyond 30B parameters within
one model family~\citep{zheng2026foreagent}. Inside a search, a predicted
outcome is treated like a measurement, so an error promotes a worse program,
prunes a better one, and stays uncorrected until something is executed. In
our experiments, a fast selector that is right on 56\% of pairs lowers the
final score of all three agents we test (\Cref{sec:exp-agents}).

This paper starts from a simple observation. To choose which program to keep,
an agent does not need to know how a program will run; it needs to know which
program will score higher. Only real execution adds information about that
question. A world model's prediction is computed from the task and code the
agent already has, so it can approach, but never exceed, the best decision
these inputs support, and its errors can pull the decision below it. We
therefore spend real executions on learning the decision itself. A
\emph{World Reward Model} (\wrm) reads the task and the candidate programs
and returns the probability that each would win if executed, without
predicting any part of the execution. It is trained on the outcomes of real
runs, and during search it takes the place of most of them
(\Cref{fig:teaser}c).

\Cref{sec:theory} makes this argument precise. A selector that first
predicts execution and then decides is a garbling of its inputs, so on the
same inputs it cannot beat the Bayes-optimal \wrm, however long it reasons
(\Cref{thm:dominance}). Predicting the run also adds an error that grows with
its length (\Cref{thm:decomposition}), and a predictor fit by likelihood to
whole execution logs can prefer the worse program on every instance
(\Cref{prop:mismatch}). A \wrm is trained on the decision instead, and its
accuracy gap is at most the square root of twice its excess cross-entropy on
executed winners (\Cref{prop:proper}).

We build \fcm, a \wrm for ML programs, by adapting an open typed decision
model~\citep{palmer2026kev} to 5,000 program pairs labeled by their executed
test scores (\Cref{sec:method}). On 12,438 held-out pairs from 26 tasks,
\fcm picks the better program 79.4\% of the time, compared with 54.5\% before
training and 53--62\% for general-purpose LLM judges (\Cref{fig:llm}), and
it reaches 71.5\% on pairs in which neither program was seen during training.
Its probabilities are also calibrated, and the 38\% of pairs it is most sure
about are 95.2\% correct. In AIDE and ML-Master, \fcm takes the place of execution
during search, and in \foreagent it takes the place of the LLM judge
(\Cref{sec:method-agent}). Mean task time drops by 4.4--7.5$\times$,
\foreagent's mean score rises from 0.527 to 0.607, and AIDE and ML-Master stay
within 0.01 of their original scores (\Cref{tab:agents}).

\paragraph{Contributions.}
\begin{itemize}
  \item World Reward Models, with proofs that predicting execution cannot
    beat deciding directly from the same inputs and can mislead the decision.
  \item A recipe for training a \wrm on executed program pairs and plugging
    it into search-based ML agents.
  \item \fcm, which is more accurate than general-purpose LLM judges, is
    calibrated, and makes three agents 4.4--7.5$\times$ faster at similar or
    better scores.
\end{itemize}

%% file: sections/setup.tex
\section{Choosing Before Executing}
\label{sec:setup}

An agent that has proposed several candidate programs must decide which of
them to execute. This section sets up that decision and the two kinds of
selectors we compare.

\paragraph{Selection before execution.}
A decision instance $X=(\ctx,\cands)$ consists of a context $\ctx$ and
candidates $\cands=(a_1,\ldots,a_K)$, drawn from a distribution $\cD$ that
includes the agent's proposal policy. In ML engineering, $\ctx$ contains the
task description and a data report, and each $a_k$ is a complete program.
Executing the candidates would produce outcomes
$S'=(s'_1,\ldots,s'_K)\sim P(\cdot\mid X)$, such as training logs, and rewards
$r_k=R(\ctx,a_k,s'_k)\in[0,1]$, such as normalized test scores. We write
$Y=\argmax_kr_k$ for the candidate that actually wins, with ties broken
uniformly at random. Before anything runs, a candidate is described by its
probability of winning and by its expected reward,
\begin{equation}
  \eta_k(X)=\Prob[Y=k\mid X],\qquad
  \Qv_k(X)=\E[r_k\mid X].
  \label{eq:q}
\end{equation}
Accuracy depends on the first and value on the second, and the two can
disagree. If candidate~1 always scores 0.6 while candidate~2 scores 0.61
with probability 0.51 and 0 otherwise, candidate~2 wins more often, but
candidate~1 has the higher expected reward.

\begin{definition}[Selector]
  \label{def:selector}
  A \emph{selector} is a possibly randomized map $\delta$ from $X$ to an index
  in $[K]$ that executes no candidate. Its value, accuracy, and regret are
  \begin{align}
    J(\delta)&=\E\bigl[\Qv_{\delta(X)}(X)\bigr],\qquad
    \Acc(\delta)=\Prob\bigl[\delta(X)=Y\bigr],\nonumber\\
    \Reg(\delta)&=\E\bigl[\max\nolimits_k\Qv_k(X)\bigr]-J(\delta).
    \label{eq:value-acc}
  \end{align}
\end{definition}

\paragraph{Predicting execution.}
A world model decides in two steps. It first predicts what executing the
candidates would produce, and then judges the prediction.

\begin{definition}[World-model selector]
  \label{def:wm}
  A \emph{world-model selector} $(\Kern,g)$ consists of a prediction kernel
  $\Kern(\cdot\mid X)$ over a space $\cZ$ of predicted traces and a decision
  rule $g$ from $(X,\imag)$ to $[K]$; it selects $g(X,\imag)$ with
  $\imag\sim\Kern(\cdot\mid X)$. Its decision distribution is
  $\pi_k(X)=\Prob[g(X,\imag)=k\mid X]$.
\end{definition}

The definition covers simulate-then-evaluate planning, which samples
$\hat s'_k\sim\hat P(\cdot\mid\ctx,a_k)$ and selects
$\argmax_k\hat R(\ctx,a_k,\hat s'_k)$~\citep{sutton1991dyna,ha2018worldmodels}.
It also covers an LLM that reasons about two programs before stating a
preference~\citep{zheng2026foreagent}, whose reasoning trace plays the role
of $\imag$. What these selectors share is that they consult nothing beyond
$X$.

\begin{assumption}[No new evidence]
  \label{ass:imagination}
  Given $X$, the predicted trace and the decision rule's internal randomness
  are independent of the real outcomes $S'$ and of the tie-breaking in $Y$.
\end{assumption}

The assumption holds for any model size and any amount of reasoning. It
fails only when the procedure gathers new evidence, for instance by training
a candidate on a subsample of the data or by querying the environment, and
such steps are executions.

\paragraph{Deciding directly.}
A \wrm skips the prediction and outputs the decision.

\begin{definition}[World Reward Model]
  \label{def:wrm}
  A \emph{World Reward Model} is a selector that maps $X$ directly, without
  generating any intermediate outcome, to either a distribution $w(X)$ in the
  probability simplex $\simplex$ over the $K$ candidates (a
  \emph{preference} \wrm, an estimate of $\eta$) or a score vector
  $q(X)\in\R^K$ (a \emph{value} \wrm, an estimate of $\Qv$), and selects the
  argmax. A preference \wrm is \emph{calibrated} if
  $\Prob[Y=k\mid w(X)=\pi]=\pi_k$ for all $\pi$ and $k$.
\end{definition}

The Bayes preference \wrm $\eta$ maximizes accuracy, and the Bayes value
\wrm $\Qv$ maximizes value. The definition constrains what is predicted, not
the output layer, so a generative judge fine-tuned to answer without
intermediate text is also a \wrm. Unlike a reward model for text, whose
labels come from human preferences, a \wrm learns from the rewards the world
returns when actions are executed. Our experiments train a preference \wrm
for $K=2$.

%% file: sections/theory.tex
\section{What Predicting Execution Can and Cannot Add}
\label{sec:theory}

\begin{figure*}[t]
  \centering
  \includegraphics[width=\textwidth]{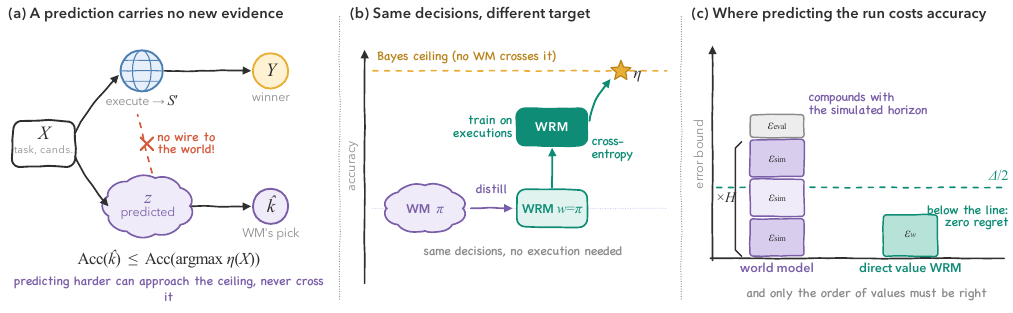}
  \caption{\textbf{Predicting execution adds no information about the
    winner.} (a)~The predicted trace $\imag$ is computed from $X$, so no
    world-model selector exceeds $\argmax\eta$ (\Cref{thm:dominance}).
    (b)~A world model's decisions are those of a \wrm with $w=\pi$; distilling
    it copies its accuracy, while training on executed outcomes targets the
    ceiling (\Cref{prop:reduction,prop:proper}). (c)~For value scores,
    predicting the run adds an error term that grows with the horizon $H$
    (\Cref{thm:decomposition}).}
  \label{fig:theory}
\end{figure*}

We now ask what a world model's prediction can contribute to the decision
(\Cref{fig:theory}). The arguments rely on standard tools (garbling, Gibbs' and Pinsker's
inequalities, the simulation lemma), and \Cref{app:proofs} gives the full
proofs.

\subsection{A prediction carries no new evidence}
\label{sec:theory-dominance}

Whatever a world model predicts is a function of what it was given, so its
decision can at best reproduce the best decision that the inputs support.

\begin{theorem}[Fixed-information ceiling]
  \label{thm:dominance}
  Under \Cref{ass:imagination}, every world-model selector $(\Kern,g)$ with
  decision distribution $\pi$ satisfies
  \begin{align}
    \Acc(\delta_{\Kern,g})&=\E\Bigl[\sum_k\pi_k(X)\eta_k(X)\Bigr]
      \le\E\bigl[\max\nolimits_k\eta_k(X)\bigr],\nonumber\\
    J(\delta_{\Kern,g})&=\E\Bigl[\sum_k\pi_k(X)\Qv_k(X)\Bigr]
      \le\E\bigl[\max\nolimits_k\Qv_k(X)\bigr].
    \label{eq:dominance}
  \end{align}
  The right-hand sides are attained by the Bayes preference \wrm $\argmax\eta$
  and the Bayes value \wrm $\argmax\Qv$. Each gap equals the
  $\pi$-weighted shortfall, \eg
  $\E\bigl[\sum_k\pi_k(X)(\max_j\Qv_j(X)-\Qv_k(X))\bigr]$ for value, and
  vanishes if and only if $\pi$ puts all its mass on maximizers for
  $\cD$-almost every $X$.
\end{theorem}

Given $X$, the decision is independent of the realized rewards. Its expected
reward is therefore a $\pi$-weighted average of the $\Qv_k$, and its chance
of being right is a $\pi$-weighted average of the $\eta_k$
(\Cref{app:proof-dominance}). In the terms of
\citet{blackwell1953equivalent}, a predicted trace is a garbling of the
inputs.

\begin{corollary}[No prediction budget crosses the ceiling]
  \label{cor:budget}
  For any sequence of world-model selectors satisfying
  \Cref{ass:imagination}, indexed by any measure of effort (rollouts,
  reasoning tokens, parameters), accuracy and value stay below the bounds in
  \cref{eq:dominance}.
\end{corollary}

\paragraph{What the ceiling does not say.}
\Cref{thm:dominance} compares decision rules that see the same information,
not networks of the same size. For a finite selector, the distance to the
ceiling is an approximation error. Reasoning spends inference-time computation
on it, which can help~\citep{chae2025wma}, while a \wrm spends real executions
on it at training time, and \Cref{prop:proper} ties that effort to accuracy.
Which route gets closer for ML programs is an empirical question
(\Cref{sec:experiments}).

\subsection{Same decisions, different targets}
\label{sec:theory-reduction}

\begin{proposition}[Behavioral equivalence and distillation]
  \label{prop:reduction}
  (i)~For every world-model selector, the preference \wrm $w=\pi$, sampled,
  reproduces its decision distribution on every instance, with equal accuracy
  and value. (ii)~Conversely, every preference \wrm $w$ is reproduced by the
  world-model selector that draws $\imag\sim w(X)$ and outputs $\imag$.
  (iii)~Over all preference \wrms, $w=\pi$ minimizes the distillation loss
  $\E\bigl[-\log w_{g(X,\imag)}(X)\bigr]$.
\end{proposition}

World-model selectors and \wrms can thus express the same decision rules.
They differ in how a rule is computed and in what it is trained to match.
The equivalence concerns which rules exist, not what they cost; it does not
imply that a network of a given size computes $\pi$ in one pass, and part~(i)
matches $\pi$ by sampling, whereas a deployed selector takes the argmax.
Distilling a world model copies its $\pi$, mistakes included. The
cross-entropy on executed winners targets $\eta$ instead, and this target
controls accuracy.

\begin{proposition}[Cross-entropy controls selection accuracy]
  \label{prop:proper}
  Let $w$ be a preference \wrm with excess cross-entropy
  $\varepsilon=\E[-\log w_Y(X)]-\E[-\log\eta_Y(X)]$. Then
  \begin{equation}
    \Acc(\delta_\eta)-\Acc(\delta_w)\le\sqrt{2\varepsilon}.
    \label{eq:proper}
  \end{equation}
\end{proposition}

The proof applies Pinsker's inequality to
$\varepsilon=\E[\mathrm{KL}(\eta\,\|\,w)]$ (\Cref{app:proof-proper}). The
bound holds at the population level and makes no claim about sample size,
but it shows that driving the training objective of \Cref{sec:method} to zero
drives the selector to the ceiling of \Cref{thm:dominance}.

\subsection{Where predicting the run costs accuracy}
\label{sec:theory-errors}

Executing a program unfolds in stages, such as preprocessing, training
epochs, and evaluation. Let the real outcome end an $H$-stage chain
$s_0=(\ctx,a)$, $s_h\sim P_h(\cdot\mid s_{h-1})$, $s'=s_H$. A world model
imitates the chain with kernels $\hat P_h$ and scores the predicted end state
with an evaluator $\hat R$, giving
$\hat q(\ctx,a)=\E_{\hat P}[\hat R(\ctx,a,\hat s_H)]$. Let
$\varepsilon_{\mathrm{sim}}=\max_h\sup_s\TV\bigl(\hat P_h(\cdot\mid s),P_h(\cdot\mid s)\bigr)$,
where $\TV(\mu,\nu)=\sup_E\abs{\mu(E)-\nu(E)}$ is the total-variation
distance, and $\varepsilon_{\mathrm{eval}}=\sup\abs{\hat R-R}$.

\begin{theorem}[Error decomposition for value scores]
  \label{thm:decomposition}
  For every $(\ctx,a)$,
  $\abs{\hat q(\ctx,a)-\Qv(\ctx,a)}\le\min\{1,H\varepsilon_{\mathrm{sim}}\}+\varepsilon_{\mathrm{eval}}$.
  For value scores with $\max_k\abs{q_k-\Qv_k(X)}\le\varepsilon$ on an
  instance, $\argmax_kq_k$ loses at most $\min\{1,2\varepsilon\}$ in value
  there, and nothing when the gap $\Delta(X)$ between the two best values
  exceeds $2\varepsilon$. Hence a simulate-then-evaluate selector has
  $\Reg\le2(\min\{1,H\varepsilon_{\mathrm{sim}}\}+\varepsilon_{\mathrm{eval}})$,
  and a direct value \wrm with sup-error $\varepsilon_w$ has
  $\Reg\le2\varepsilon_w$.
\end{theorem}

The simulation term is the one-step form of the compounding error of
model-based reinforcement learning~\citep{kearns2002near,janner2019mbpo}. The
next result shows that the gap is real and that a wrong prediction can
mislead the decision systematically rather than at random.

\begin{proposition}[Objective mismatch in a restricted family]
  \label{prop:mismatch}
  Let the outcome be a reward bit $y\sim\Bern(q_k)$ followed by $d$
  decision-irrelevant bits $\xi_j\sim\Bern(m_k)$, with $q_1>q_2$ and $m_1<m_2$,
  and let the model family tie all $1+d$ bits of a candidate to one rate
  $\theta_k$. The maximum-likelihood world model in this family, with an exact
  evaluator, selects the lower-value candidate on every instance whenever
  $d(m_2-m_1)>q_1-q_2$, whereas the same family fitted to the reward bit alone
  selects the higher-value candidate.
\end{proposition}

With $q=(0.8,0.3)$, $m=(0.1,0.9)$, and $d=10$, the likelihood fit gives
$\hat\theta\approx(0.16,0.85)$, so the world model picks the worse candidate
every time, at a regret of 0.5 (\Cref{app:proof-mismatch}). The construction
mirrors an execution log, in which a few tokens report the metric and many
more report progress; a model fit to the whole log learns what makes the log
probable rather than what makes the score high~\citep{lambert2020objective}.
In WMA, likewise, predicting the full next observation lowers success below
that obtained with no prediction at all~\citep{chae2025wma}.

Screening more candidates does not make up for such errors. In an idealized
Gaussian model (\Cref{app:screening}), a judge with pairwise accuracy 0.615
needs 907 candidates to match a judge with accuracy 0.794 that sees 8.

%% file: sections/method.tex
\section{Building and Using a World Reward Model}
\label{sec:method}

\begin{figure*}[t]
  \centering
  \includegraphics[width=\textwidth]{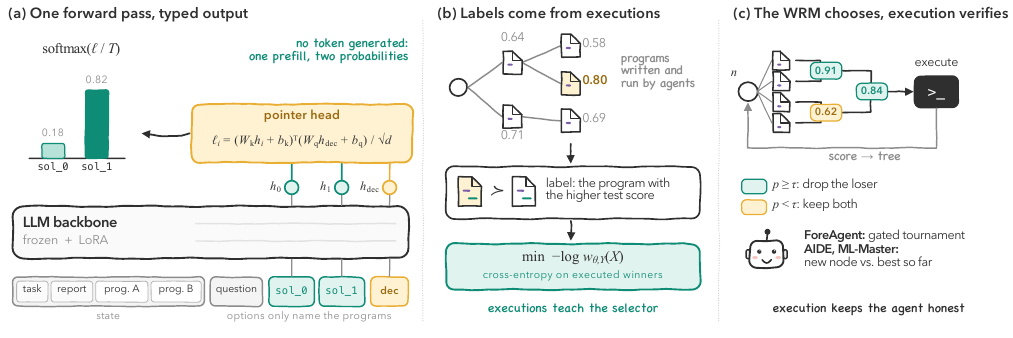}
  \caption{\textbf{A World Reward Model is a typed decision model trained on
    what the world returns.} (a)~Task, data report, and both programs form a
    shared state; a pointer head scores two short options against a decision
    token in one forward pass, with no generated token. (b)~Program pairs
    labeled by their executed test scores train the \wrm with cross-entropy.
    (c)~Inside an agent, the \wrm settles comparisons, dropping a loser only
    when confident; execution verifies the chosen program, and its score
    returns to the search tree.}
  \label{fig:method}
\end{figure*}

We now build a pairwise \wrm for ML programs, train it toward $\eta$ with the
cross-entropy of \Cref{prop:proper}, and plug it into search-based agents
(\Cref{fig:method}).

\subsection{A \wrm as a typed decision model}
\label{sec:method-model}

A selector inside an agent has to read long programs, compare them, and
return a probability that the agent can threshold. Typed decision
models~\citep{typesafe2026jev,palmer2026kev} offer this interface, and we
adopt the open Kev architecture~\citep{palmer2026kev}. A request is
serialized as one token sequence. It begins with a state holding the task
name, the task description, the data report of~\citet{zheng2026foreagent},
and both complete programs, followed by a question asking which program will
perform better, two short options, \texttt{solution\_0} and
\texttt{solution\_1}, and a decision token. A frozen decoder backbone with
LoRA adapters reads the sequence once. Let $h_{\mathrm{dec}}$ be the
last-layer state at the decision token and $h_i$ the state at the closing
token of option $i$. A pointer head computes
\begin{align}
  \ell_i&=(W_kh_i+b_k)^{\transpose}(W_qh_{\mathrm{dec}}+b_q)/\sqrt{d},\nonumber\\
  w_i(X)&=\softmax(\ell/T)_i,
  \label{eq:pointer}
\end{align}
where the affine projections map to $d=256$ dimensions and the temperature
$T$ is fitted after training. Because the decision token attends to both
programs, the judgment is comparative. Every output is a valid choice with a
probability, and a decision costs one forward pass over the input with no
decoding.

\subsection{Training on what the world returns}
\label{sec:method-training}

\paragraph{Data.}
We use a frozen copy of the Predict-Before-Execute
corpus~\citep{zheng2026foreagent}, which contains 18,438 pairs of complete
programs written by ML agents for 26 MLE-bench tasks. Each pair is labeled
by which program achieved the better test score when executed, using the
corpus's metric directions and its filtering of near-ties. We split the
source pairs into 5,000 for training, 250 for development, 250 for
calibration, 500 for an internal test, and 12,438 for the final test, and we
keep identical program pairs within one partition. All 26 tasks appear in
training, and a program can recur across pairs.

\paragraph{Objective and recipe.}
The \wrm minimizes the cross-entropy of the executed winner,
\begin{equation}
  \cL(\theta)=\E_{(X,Y)}\bigl[-\log w_{\theta,Y}(X)\bigr],
  \label{eq:loss}
\end{equation}
whose population minimizer is $\eta$. \fcm starts from the released Kev-9B,
a frozen Qwen3.5-9B-Base backbone with a LoRA adapter and a pointer head
trained for general typed decisions, and updates only the adapter and the
head. Each epoch shows every pair in one of six variants (three input
formats, two candidate orders). Four epochs, or 628 updates, take 4.2 hours
on eight A800 GPUs. We select the checkpoint by development accuracy and fit
$T$ by NLL on the calibration partition, which gives
$T=1.086$~\citep{guo2017calibration}. Final-test labels are used for neither
step (\Cref{app:details}).

\subsection{Plugging a \wrm into search-based agents}
\label{sec:method-agent}

\begin{algorithm}[t]
  \caption{Tree search with a \wrm. Colored lines replace the execution of
    each new candidate.}
  \label{alg:search}
  \begin{algorithmic}[1]
    \REQUIRE task context $\ctx$, \wrm $w$, budget $B$
    \STATE $\mathcal T\leftarrow$ initial drafts;\quad $a^\star\leftarrow$ a draft
    \FOR{$t=1,\ldots,B$}
      \STATE $n\leftarrow\textsc{Select}(\mathcal T)$ \COMMENT{greedy or UCT}
      \STATE $a\leftarrow$ LLM improves, debugs, or redrafts $n$
      \STATE \textcolor{papergreen}{$p\leftarrow w\bigl(\ctx,(a,a^\star)\bigr)_a$}
        \COMMENT{was: execute $a$}
      \STATE \textcolor{papergreen}{$\textsc{Update}(\mathcal T,a,p)$}
        \COMMENT{node score or reward}
      \STATE $a^\star\leftarrow$ highest-scoring node of $\mathcal T$
    \ENDFOR
    \STATE execute $a^\star$, repair it if needed, and submit
  \end{algorithmic}
\end{algorithm}

\paragraph{How search-based agents work.}
Most ML-engineering agents organize their work as a tree whose nodes are
candidate
programs~\citep{jiang2025aide,liu2025mlmaster,toledo2025aira}. In each step a
selection policy picks a node, an LLM writes a child program by improving the
selected one, fixing a bug, or drafting anew, and the child is executed to
obtain its validation score. The score serves two purposes. It tells the
agent whether the child beats the best program so far, and it steers the
policy toward promising parts of the tree. When the budget runs out, the
agent submits the best program it has found.

\paragraph{Where the \wrm enters.}
A \wrm takes over the scoring role of execution and leaves the rest of the
agent untouched. Instead of running a new child, the agent asks the \wrm how
likely the child is to beat the incumbent and uses this probability where it
used the score (\Cref{alg:search}). Only the submitted program is executed,
with repairs if that run fails.

\paragraph{Two examples.}
AIDE~\citep{jiang2025aide} runs a greedy tree search. It drafts several
initial solutions, repeatedly improves the best node or debugs a failed one,
and keeps the node with the highest validation score. With a \wrm, each new
node is compared with the incumbent, and the log-odds of winning are added
to the node scores that the greedy policy reads.
ML-Master~\citep{liu2025mlmaster} runs a search in the style of Monte Carlo
tree search, which chooses nodes by an upper confidence bound and backs
rewards up the tree. With a \wrm, the reward of a new node is its probability
of beating the best solution so far, and this probability is backed up in
place of the executed score. In both agents the search then executes no
candidate before the final submission. \foreagent~\citep{zheng2026foreagent}
already avoids most executions. An LLM judges a batch of proposals in a
pairwise tournament, and only the top-ranked program runs. There \fcm takes
the place of the LLM judge and eliminates a loser only when the winner's
probability reaches $\tau=0.7$ (\Cref{alg:tournament}).

%% file: sections/experiments.tex
\section{Experiments}
\label{sec:experiments}

We evaluate \fcm with four questions. \textbf{Q1:} Does learning from real
execution make a selector accurate, including on programs it has never seen
(\Cref{sec:exp-main})? \textbf{Q2:} How does it compare with general-purpose
LLM judges (\Cref{sec:exp-accuracy})? \textbf{Q3:} Are its probabilities
reliable enough to act on (\Cref{sec:exp-probabilities})? \textbf{Q4:} Does it
make ML agents faster without hurting their results
(\Cref{sec:exp-agents})?

\subsection{Setup}
\label{sec:exp-setup}

\paragraph{Data and baselines.}
Unless stated otherwise, results are on the final test, 12,438 program pairs
held out from training. The main comparison sets \fcm against its own
initialization, the released Kev-9B, with the same inputs, labels, and
architecture, so that the difference isolates what learning from execution
adds. The general-purpose judges are Qwen3.5-9B with and without reasoning
(1,024-token budget), evaluated on 9,879 final-test pairs, the typed decision
model Jev~\citep{typesafe2026jev}, evaluated on the full final test, and
DeepSeek-V3.2 and GPT-5.1, whose accuracies \citet{zheng2026foreagent} report
on the full corpus, of which our final test makes up two thirds.

\paragraph{Metrics.}
We report pairwise accuracy, task-macro accuracy, negative log-likelihood
(NLL), the Brier score summed over both classes, and the expected calibration
error (ECE) over ten equal-width bins. We call a difference significant when
a paired bootstrap, over pairs or over tasks where stated, rejects zero at
the 5\% level. Since every program belongs to a single task, resampling whole
tasks also accounts for pairs that share a program, which makes it the
conservative choice.

\subsection{Q1: Learning from execution makes the selector accurate}
\label{sec:exp-main}

\begin{table}[t]
  \caption{\textbf{Training on execution outcomes lifts a typed selector by
    about 24 points, including on unseen programs.} Pairwise accuracy (\%) of
    the released Kev-9B and of \fcm, its trained version, under identical
    inputs and labels; ``unseen'' programs are absent from training.
    $\Delta$: paired difference, significant in every row under a paired
    bootstrap (over tasks for the macro row).}
  \label{tab:main}
  \centering
  \footnotesize
  \setlength{\tabcolsep}{2.6pt}
  \begin{tabular}{lrrl}
    \toprule
    Subset (pairs) & Kev-9B & \fcm & $\Delta$ \\
    \midrule
    Final test (12,438) & 54.53 & \textbf{79.44} & $+24.91$ \\
    \quad no data report & 54.46 & \textbf{78.55} & $+24.09$ \\
    \quad $\ge$1 unseen (5,996) & 54.10 & \textbf{77.18} & $+23.08$ \\
    \quad 2 unseen (501) & 48.10 & \textbf{71.46} & $+23.35$ \\
    Internal test (500) & 52.60 & \textbf{78.20} & $+25.60$ \\
    Task macro (26) & 54.60 & \textbf{72.69} & $+18.10$ \\
    \bottomrule
  \end{tabular}
\end{table}

Training on execution outcomes raises accuracy on the final test from
54.53\% to 79.44\%, a paired gain of 24.91 points (\Cref{tab:main}). With
every task weighted equally, the gain is 18.10 points, and it stays
significant when whole tasks are resampled, so a few large tasks do not
carry it.

Programs can appear in several pairs, so the final test mixes new pairs of
known programs with pairs of new programs. On the 501 pairs in which neither
program was seen during training, \fcm reaches 71.46\%. This is 8 points below
its overall accuracy, but Kev-9B, which was trained on none of these programs,
drops by 6.4 points on the same pairs, and the gain stays at 23.35 points.
These pairs are harder for both models, and the benefit of training does not
depend on having seen the programs. Removing the data report at test time
costs 0.89 points while shortening the input by 63.9\%.

\subsection{Q2: Comparison with general-purpose LLMs}
\label{sec:exp-accuracy}

\begin{figure}[t]
  \centering
  \includegraphics[width=\linewidth]{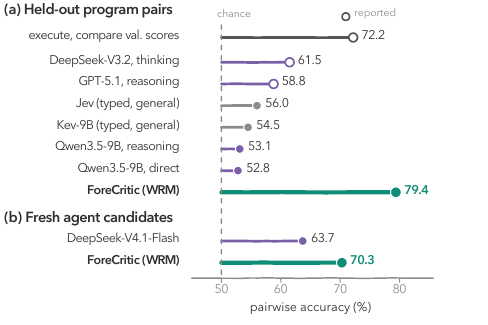}
  \caption{\textbf{A \wrm trained on real executions outperforms
    general-purpose judges.} (a)~Held-out program pairs. \fcm, Jev, and
    Kev-9B on the final test; Qwen3.5-9B on its 9,879-pair subset, where
    \fcm scores 79.9\%. Hollow markers are reported by
    \citet{zheng2026foreagent} on the full corpus, including how often the
    executed validation score picks the test-score winner. (b)~182 judgments
    of fresh programs that an agent wrote on ten MLE-Dojo tasks.}
  \label{fig:llm}
\end{figure}

Large reasoning models reach 61.5\% (DeepSeek-V3.2) and 58.8\% (GPT-5.1) on
the full corpus~\citep{zheng2026foreagent}, general typed decision models stay
close to chance (Jev 56.0\%, Kev-9B 54.5\%), and the same-size Qwen3.5-9B
reaches 53.1\% with reasoning and 52.8\% without (\Cref{fig:llm}a). On the
9,879 pairs shared with the Qwen evaluation, \fcm is right 79.9\% of the time,
a significant gap of 26.8 points. Within one model, reasoning moves the
generative judge by 0.3 points, whereas training on executions moves the typed
judge by 25.3 points. This is the pattern \Cref{cor:budget} predicts, since
thinking harder about the run does not supply the missing evidence and real
runs do.

Execution itself is a useful reference. Ranking two executed programs by
their validation scores picks the test-score winner on 72.2\% of the corpus's
pairs~\citep{zheng2026foreagent}, and \fcm, which learns from test outcomes, does
better without running either program, on an overlapping set of pairs.

The advantage carries over to programs from another source. On 91 pairs of
programs that \foreagent executed on ten MLE-Dojo
tasks~\citep{qiang2025mledojo}, each judged in both orders, \fcm is right on
70.3\% of judgments and DeepSeek-V4.1-Flash on 63.7\% (\Cref{fig:llm}b), a
difference this small sample does not make significant (\Cref{app:transfer}).
DeepSeek also read 1.44M input tokens and wrote 2,987, while \fcm generates
none.

\subsection{Q3: The probabilities are calibrated and informative}
\label{sec:exp-probabilities}

\begin{figure}[t]
  \centering
  \includegraphics[width=\linewidth]{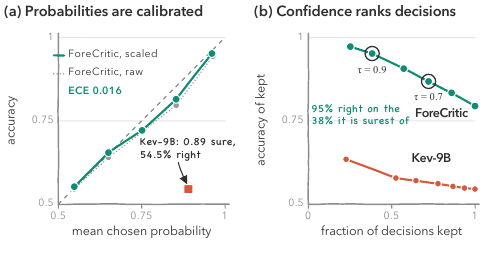}
  \caption{\textbf{\fcm knows when it is right.} Final test, 12,438 pairs.
    (a)~Mean chosen probability against accuracy per bin; Kev-9B as one point
    (mean confidence, accuracy). (b)~Accuracy of the decisions kept at
    threshold $\tau$ against the fraction kept.}
  \label{fig:probabilities}
\end{figure}

An agent can act on a probability only if it means what it says. Before any
rescaling, \fcm's ECE is 0.025, while Kev-9B reports an average confidence of
0.89 at 54.5\% accuracy, an ECE of 0.345 (\Cref{fig:probabilities}a).
Temperature scaling lowers \fcm's NLL from 0.4364 to 0.4332 and its ECE to
0.016 without changing any decision. It also repairs Kev-9B's ECE, but not
its accuracy (\Cref{fig:probquality}). The probabilities rank decisions as
well (\Cref{fig:probabilities}b). Keeping only decisions above 0.7 retains
72.2\% of pairs at 86.8\% accuracy, and a threshold of 0.9 retains 38.3\% at
95.2\%, whereas Kev-9B's most confident 22.7\% are 63.5\% correct.

\subsection{Q4: The \wrm inside three ML agents}
\label{sec:exp-agents}

\begin{table}[t]
  \caption{\textbf{An accurate \wrm keeps or raises agent scores at a
    fraction of the time; a near-chance one lowers them.} Mean native score
    and mean task time over ten MLE-Dojo tasks with each agent's original
    selector, with Jev, and with \fcm, under the same agent configuration;
    one scored run per task, timed from start to submission. Native metrics
    differ in scale and direction (7 of 10 are higher-is-better;
    \Cref{fig:pertask}).}
  \label{tab:agents}
  \centering
  \footnotesize
  \setlength{\tabcolsep}{2.5pt}
  \begin{tabular}{lrrrrrr}
    \toprule
     & \multicolumn{3}{c}{Native score} & \multicolumn{3}{c}{Task time (s) $\downarrow$} \\
    \cmidrule(lr){2-4}\cmidrule(lr){5-7}
    Agent & Orig. & Jev & \fcm & Orig. & Jev & \fcm \\
    \midrule
    AIDE & 0.600 & 0.499 & 0.599 & 3,164 & 738 & \textbf{725} \\
    ML-Master & 0.616 & 0.484 & 0.609 & 3,711 & 839 & \textbf{497} \\
    \foreagent & 0.527 & 0.472 & \textbf{0.607} & 4,111 & 2,094 & \textbf{884} \\
    \bottomrule
  \end{tabular}
\end{table}

\Cref{tab:agents} compares each agent with its original selector, with Jev,
and with \fcm under the same configuration. In AIDE and ML-Master, \fcm
takes the place of execution during search, so only the final program and
its repairs run. Mean task time falls from 3,164 to 725 seconds in AIDE
($4.4\times$) and from 3,711 to 497 seconds in ML-Master ($7.5\times$), and
the mean scores stay within 0.01 of the original ones. In \foreagent, \fcm
takes the place of the LLM judge, mean task time falls from 4,111 to 884
seconds ($4.7\times$), and the mean score rises from 0.527 to 0.607. \fcm's
446 comparisons take under 3\% of the task time, and training it once took 33
A800 GPU-hours (\Cref{app:agents}).

Accuracy is what makes these savings safe. Jev is also a fast typed
selector, but it is right on only 56.0\% of pairs, and with Jev all three
agents end with lower scores (0.499, 0.484, and 0.472). A selector that is
often wrong misleads the search, as \Cref{sec:theory-errors} describes for
wrong predictions. Each entry averages one run per task, so we read these
comparisons as descriptive (per-task results in \Cref{fig:pertask}).

%% file: sections/related-work.tex
\section{Related Work}
\label{sec:related-work}

\paragraph{World models for planning and for LLM agents.}
Planning with a learned model of the environment goes back to
Dyna~\citep{sutton1991dyna} and continues through World
Models~\citep{ha2018worldmodels}, MuZero~\citep{schrittwieser2020muzero}, and
Dreamer~\citep{hafner2023dreamerv3}; \citet{ding2025survey} survey the area.
LLMs now act as world models for reasoning~\citep{hao2023rap}, web
navigation~\citep{gu2024webdreamer,chae2025wma}, text
games~\citep{wang2024simulators}, code
execution~\citep{li2025codeio,copet2025cwm}, and ML
engineering~\citep{zheng2026foreagent}, whose problem and corpus we adopt. We
show what such predictions cannot add and train the judge on real executions
instead.

\paragraph{Learning values instead of models.}
Q-learning~\citep{watkins1992q} learns action values without a model.
Value-aware model learning~\citep{farahmand2017vaml} and value
equivalence~\citep{grimm2020value} ask a model to be accurate only where values
depend on it, and \citet{lambert2020objective} show that models trained by
likelihood can hurt control. We carry these ideas to one-step selection among
programs.

\paragraph{Reward models, verifiers, and judges.}
Reward models score outputs by human
preference~\citep{christiano2017deep,ouyang2022training}, verifiers check
solutions~\citep{cobbe2021verifiers,lightman2024verify}, LLM judges reason in
text~\citep{zheng2023judging,zhang2025genrm}, and best-of-$N$ selection with
such scorers is a common use of test-time
compute~\citep{brown2024monkeys,snell2025scaling}. A \wrm instead predicts,
before an action runs, the reward the world will return, and learns it from
executed outcomes.

\paragraph{Predicting program performance.}
NAS performance predictors~\citep{white2021powerful}, learning-curve
extrapolation~\citep{domhan2015speeding}, and regression language models for
code~\citep{akhauri2026regression} estimate quality without full training,
usually for one program at a time; a \wrm compares free-form programs for the
same task, inside agents that iterate with execution
feedback~\citep{jiang2025aide,liu2025mlmaster,ou2025automind,yang2025rdagent,toledo2025aira}.
Typed decision models~\citep{typesafe2026jev,palmer2026kev} provide its
interface but, as \Cref{tab:main} shows, need training on execution to work
for ML programs.

%% file: sections/conclusion.tex
\section{Limitations}
\label{sec:limitations}

World Reward Models invite a larger agenda than one paper can complete. The
selector here learns from executions on a family of ML tasks and settles one
comparison at a time. The general \wrm we have in mind would carry over to new
tasks and domains, rank many candidates in one pass, and keep learning from
its agent's executions, so that every run both uses and improves it. We would
also have liked to compare training signals and decision interfaces at
matched scale, to extend the theory from Bayes-optimal decisions to finite
models and data, and to judge selectors by final reward under a fixed budget
across many agents, with value \wrms that predict how much a candidate will
score. The same recipe applies wherever actions are expensive to try and
their outcomes can be recorded, from experimental design to compiler tuning
and robotics, and we leave these domains to future work.

\section{Conclusion}
\label{sec:conclusion}

ML agents search by executing programs, and execution is both their most
reliable signal and their main cost. Predicting execution saves runs but adds
no evidence about which program wins, and its errors mislead the search. A
World Reward Model instead learns from real executions the probability that
one program beats another and leaves execution to verify the program the
agent keeps. \fcm picks the better of two ML programs 79.4\% of the time,
well above general-purpose LLM judges, and makes three agents
4.4--7.5$\times$ faster at similar or better scores.

%% file: sections/impact.tex
This paper studies how agents choose which actions to execute. World Reward
Models make that choice cheaper, which can lower the compute and energy spent
on executing poor candidates in ML-engineering agents. The same efficiency
makes autonomous agents cheaper to run in other settings, including ones where
actions have consequences outside a sandbox; a confidently wrong selector can
steer an agent toward a harmful action without the friction that execution
feedback provides. We recommend keeping execution as verification and using a
selector's calibrated probabilities to route uncertain decisions to execution
or to a human.

%% file: sections/acknowledgements.tex

%% file: appendix.tex
\section{Proofs}
\label{app:proofs}

Throughout
\Cref{app:proof-dominance,app:proof-reduction,app:proof-proper,app:proof-decomposition,app:proof-mismatch},
$X\sim\cD$, rewards lie in $[0,1]$, conditional statements hold for
$\cD$-almost every $X$, and $\Ind\{\cdot\}$ is the indicator of an event. We
use one property of the total-variation distance
$\TV(\mu,\nu)=\sup_E\abs{\mu(E)-\nu(E)}$: for probability measures $\mu,\nu$
and a measurable $f$ with values in $[0,1]$,
\begin{align}
  \E_\mu f-\E_\nu f
  &=\int_0^1\bigl(\mu(f>t)-\nu(f>t)\bigr)\,dt
  &&\text{(layer-cake representation of $f\in[0,1]$)}\nonumber\\
  &\le\int_0^1\TV(\mu,\nu)\,dt=\TV(\mu,\nu),
  \label{eq:tv-test}
\end{align}
and exchanging $\mu$ and $\nu$ gives $\abs{\E_\mu f-\E_\nu f}\le\TV(\mu,\nu)$.

\subsection{Fixed-information ceiling}
\label{app:proof-dominance}

\begin{proof}[Proof of \Cref{thm:dominance}]
  Fix $X$ and write $\hat k=g(X,\imag)$ for the selected index. By
  \Cref{def:wm}, $\hat k$ is a function of $X$, of $\imag\sim\Kern(\cdot\mid X)$,
  and of the rule's internal randomness. By \Cref{ass:imagination}, the last
  two are independent of the outcomes and of the tie-breaking in $Y$ given
  $X$, so
  \begin{equation}
    \hat k\indep(S',Y)\mid X.
    \label{eq:ci}
  \end{equation}

  \emph{Value.} Exactly one index is selected, so
  $r_{\hat k}=\sum_k r_k\Ind\{\hat k=k\}$, and
  \begin{align*}
    \E\bigl[r_{\hat k}\mid X\bigr]
    &=\sum_{k=1}^K\E\bigl[r_k\Ind\{\hat k=k\}\mid X\bigr]
    &&\text{(linearity)}\\
    &=\sum_{k=1}^K\E\bigl[r_k\mid X\bigr]\,\Prob\bigl[\hat k=k\mid X\bigr]
    &&\text{(\cref{eq:ci}; $r_k=R(\ctx,a_k,s'_k)$ depends on $X$ and $S'$ only)}\\
    &=\sum_{k=1}^K\pi_k(X)\,\Qv_k(X)
    &&\text{(definitions of $\pi$ and $\Qv$)}\\
    &\le\Bigl(\max_j\Qv_j(X)\Bigr)\sum_{k=1}^K\pi_k(X)
    &&\text{($\pi_k(X)\ge0$)}\\
    &=\max_j\Qv_j(X)
    &&\text{($\textstyle\sum_k\pi_k(X)=1$)}.
  \end{align*}

  \emph{Accuracy.} The events $\{\hat k=k,\ Y=k\}$, $k\in[K]$, partition
  $\{\hat k=Y\}$, so
  \begin{align*}
    \Prob\bigl[\hat k=Y\mid X\bigr]
    &=\sum_{k=1}^K\Prob\bigl[\hat k=k,\ Y=k\mid X\bigr]\\
    &=\sum_{k=1}^K\Prob\bigl[\hat k=k\mid X\bigr]\,\Prob\bigl[Y=k\mid X\bigr]
    &&\text{(\cref{eq:ci})}\\
    &=\sum_{k=1}^K\pi_k(X)\,\eta_k(X)
    \;\le\;\max_j\eta_j(X)
    &&\text{(as for the value)}.
  \end{align*}
  By the tower property, $J(\delta_{\Kern,g})=\E\bigl[\E[r_{\hat k}\mid X]\bigr]$
  and $\Acc(\delta_{\Kern,g})=\E\bigl[\Prob[\hat k=Y\mid X]\bigr]$; taking
  expectations of the two displays gives \cref{eq:dominance}.

  \emph{Gap.} Since $\sum_k\pi_k(X)=1$,
  \begin{align*}
    \E\Bigl[\max_j\Qv_j(X)\Bigr]-J(\delta_{\Kern,g})
    &=\E\Bigl[\sum_{k=1}^K\pi_k(X)\max_j\Qv_j(X)-\sum_{k=1}^K\pi_k(X)\Qv_k(X)\Bigr]\\
    &=\E\Bigl[\sum_{k=1}^K\pi_k(X)\bigl(\max_j\Qv_j(X)-\Qv_k(X)\bigr)\Bigr],
  \end{align*}
  and the same computation with $\eta$ in place of $\Qv$ gives the accuracy
  gap. Each summand is a product of two nonnegative factors, so a gap is zero
  if and only if, for $\cD$-almost every $X$, $\pi_k(X)>0$ implies that $k$
  maximizes $\Qv(X)$ (respectively $\eta(X)$).

  \emph{Attainment.} The rule $\delta^\star(X)=\argmax_k\Qv_k(X)$, with ties
  broken by the smallest index, depends on $X$ only and generates no outcome,
  so it is a value \wrm. Its decision distribution is a point mass on a
  maximizer, so the gap vanishes and $J(\delta^\star)=\E[\max_k\Qv_k(X)]$.
  The same argument shows that $\argmax_k\eta_k$ is a preference \wrm with
  $\Acc=\E[\max_k\eta_k(X)]$.
\end{proof}

\begin{proof}[Proof of \Cref{cor:budget}]
  Let $(\Kern_b,g_b)_{b\in B}$ be world-model selectors indexed by any effort
  parameter $b$, each satisfying \Cref{ass:imagination}. \Cref{thm:dominance}
  applies to each $b$, and its right-hand sides do not depend on $b$, so
  \begin{equation*}
    \sup_{b\in B}\Acc(\delta_{\Kern_b,g_b})\le\E\Bigl[\max_k\eta_k(X)\Bigr],
    \qquad
    \sup_{b\in B}J(\delta_{\Kern_b,g_b})\le\E\Bigl[\max_k\Qv_k(X)\Bigr],
  \end{equation*}
  and every limit of accuracies or values along the family obeys the same
  bounds.
\end{proof}

\subsection{Behavioral equivalence and distillation}
\label{app:proof-reduction}

\begin{proof}[Proof of \Cref{prop:reduction}]
  \emph{(i) From a world model to a \wrm.} Fix a world-model selector
  $(\Kern,g)$ and set $w=\pi$; this is a map from $X$ to $\simplex$ that
  generates no outcome. The sampled \wrm draws $\tilde k\sim\pi(X)$ with fresh
  randomness, independent of $(S',Y)$ given $X$. For every $k$,
  \begin{equation*}
    \Prob\bigl[\tilde k=k\mid X\bigr]=\pi_k(X)=\Prob\bigl[g(X,\imag)=k\mid X\bigr],
  \end{equation*}
  so the two selectors have the same decision distribution on every instance.
  Both satisfy \cref{eq:ci}, so the computation in the proof of
  \Cref{thm:dominance} applies to each:
  \begin{equation*}
    \E\bigl[r_{\tilde k}\mid X\bigr]=\sum_k\pi_k(X)\Qv_k(X)=\E\bigl[r_{\hat k}\mid X\bigr],
    \qquad
    \Prob\bigl[\tilde k=Y\mid X\bigr]=\sum_k\pi_k(X)\eta_k(X)=\Prob\bigl[\hat k=Y\mid X\bigr].
  \end{equation*}
  Taking expectations gives equal value and equal accuracy.

  \emph{(ii) From a \wrm to a world model.} Given a preference \wrm $w$, take
  $\cZ=[K]$, $\Kern(\cdot\mid X)=w(X)$, and $g(X,\imag)=\imag$. The trace is
  drawn with randomness independent of $(S',Y)$ given $X$, so
  \Cref{ass:imagination} holds, and
  $\Prob[g(X,\imag)=k\mid X]=\Kern(\{k\}\mid X)=w_k(X)$.

  \emph{(iii) Distillation.} Fix $X$ and abbreviate $\pi_k=\pi_k(X)$ and
  $w_k=w_k(X)$. The teacher outputs $k$ with probability $\pi_k$, so
  \begin{align*}
    \E\bigl[-\log w_{g(X,\imag)}(X)\mid X\bigr]
    &=-\sum_{k=1}^K\pi_k\log w_k\\
    &=-\sum_{k=1}^K\pi_k\log\pi_k+\sum_{k=1}^K\pi_k\log\frac{\pi_k}{w_k}
    &&\text{(add and subtract $\textstyle\sum_k\pi_k\log\pi_k$)}\\
    &=H(\pi)+\mathrm{KL}(\pi\,\|\,w)\\
    &\ge H(\pi)
    &&\text{(Gibbs' inequality)},
  \end{align*}
  with equality if and only if $w=\pi$. The entropy $H(\pi)$ does not depend
  on $w$, so the conditional loss is minimized exactly at $w(X)=\pi(X)$, and
  integrating over $X$ shows that $w=\pi$ minimizes the distillation loss,
  uniquely up to $\cD$-null sets.

  \emph{Sampling versus argmax.} Part~(i) matches the teacher by sampling. The
  deterministic rule $\argmax_kw_k$ with $w=\pi$ can be less or more accurate:
  with $K=2$ and $\eta=(0.9,0.1)$, the teacher $\pi=(0.4,0.6)$ is right with
  probability $0.4\cdot0.9+0.6\cdot0.1=0.42$ and its argmax with probability
  $0.1$, whereas for $\pi=(0.6,0.4)$ the teacher is right with probability
  $0.58$ and its argmax with probability $0.9$. Nor need $\pi$ equal $\eta$,
  so distillation inherits the teacher's errors.
\end{proof}

\subsection{Cross-entropy controls selection accuracy}
\label{app:proof-proper}

\begin{proof}[Proof of \Cref{prop:proper}]
  \emph{Step 1: the excess cross-entropy is an expected KL divergence.} Given
  $X$, $Y$ has law $\eta(X)$, so
  \begin{align*}
    \varepsilon
    &=\E\bigl[-\log w_Y(X)\bigr]-\E\bigl[-\log\eta_Y(X)\bigr]
     =\E\Bigl[\E\Bigl[\log\frac{\eta_Y(X)}{w_Y(X)}\Bigm|X\Bigr]\Bigr]\\
    &=\E\Bigl[\sum_{k=1}^K\eta_k(X)\log\frac{\eta_k(X)}{w_k(X)}\Bigr]
     =\E\bigl[\mathrm{KL}\bigl(\eta(X)\,\|\,w(X)\bigr)\bigr].
  \end{align*}

  \emph{Step 2: accuracy of a deterministic rule.} For a selector $\delta$
  that is a function of $X$,
  $\Acc(\delta)=\E\bigl[\Prob[Y=\delta(X)\mid X]\bigr]=\E\bigl[\eta_{\delta(X)}(X)\bigr]$.
  With $i(X)=\argmax_k\eta_k(X)$ and $j(X)=\argmax_kw_k(X)$,
  \begin{equation*}
    \Acc(\delta_\eta)-\Acc(\delta_w)=\E\bigl[\eta_{i(X)}(X)-\eta_{j(X)}(X)\bigr].
  \end{equation*}

  \emph{Step 3: the gap on one instance.} Fix $X$ and drop it from the
  notation. If $i=j$, the gap is zero. Otherwise $w_i\le w_j$ by the choice of
  $j$, and
  \begin{align*}
    \eta_i-\eta_j
    &=(\eta_i-w_i)+(w_i-w_j)+(w_j-\eta_j)\\
    &\le(\eta_i-w_i)+(w_j-\eta_j)
    &&\text{($w_i-w_j\le0$)}\\
    &\le\abs{\eta_i-w_i}+\abs{\eta_j-w_j}\;\le\;\norm{\eta-w}_1\\
    &\le\sqrt{2\,\mathrm{KL}(\eta\,\|\,w)}
    &&\text{(Pinsker's inequality)}.
  \end{align*}

  \emph{Step 4: averaging.} Taking expectations and applying Jensen's
  inequality to the concave square root,
  \begin{equation*}
    \Acc(\delta_\eta)-\Acc(\delta_w)
    \le\E\Bigl[\sqrt{2\,\mathrm{KL}\bigl(\eta(X)\,\|\,w(X)\bigr)}\Bigr]
    \le\sqrt{2\,\E\bigl[\mathrm{KL}\bigl(\eta(X)\,\|\,w(X)\bigr)\bigr]}
    =\sqrt{2\varepsilon}.
    \qedhere
  \end{equation*}
\end{proof}

\subsection{Error decomposition}
\label{app:proof-decomposition}

\begin{lemma}[Composition of simulation errors]
  \label{lem:tv-composition}
  Let $\mu_h$ and $\hat\mu_h$ be the laws of $s_h$ under the real kernels and
  under the model kernels, started from the same $s_0$. Then
  $\TV(\hat\mu_H,\mu_H)\le\min\{1,H\varepsilon_{\mathrm{sim}}\}$.
\end{lemma}

\begin{proof}
  For $h\ge1$ and every measurable event $E$,
  $\mu_h(E)=\int P_h(E\mid s)\,\mu_{h-1}(ds)$ and
  $\hat\mu_h(E)=\int\hat P_h(E\mid s)\,\hat\mu_{h-1}(ds)$. Adding and
  subtracting $\int P_h(E\mid s)\,\hat\mu_{h-1}(ds)$,
  \begin{equation*}
    \hat\mu_h(E)-\mu_h(E)
    =\underbrace{\int\bigl(\hat P_h(E\mid s)-P_h(E\mid s)\bigr)\,\hat\mu_{h-1}(ds)}_{(\mathrm{I})}
    +\underbrace{\int P_h(E\mid s)\,\bigl(\hat\mu_{h-1}-\mu_{h-1}\bigr)(ds)}_{(\mathrm{II})}.
  \end{equation*}
  For every $s$,
  $\abs{\hat P_h(E\mid s)-P_h(E\mid s)}\le\TV\bigl(\hat P_h(\cdot\mid s),P_h(\cdot\mid s)\bigr)\le\varepsilon_{\mathrm{sim}}$,
  so $\abs{(\mathrm{I})}\le\varepsilon_{\mathrm{sim}}$. The map
  $s\mapsto P_h(E\mid s)$ takes values in $[0,1]$, so \cref{eq:tv-test} gives
  $\abs{(\mathrm{II})}\le\TV(\hat\mu_{h-1},\mu_{h-1})$. Taking the supremum
  over $E$,
  \begin{equation*}
    \TV(\hat\mu_h,\mu_h)\le\varepsilon_{\mathrm{sim}}+\TV(\hat\mu_{h-1},\mu_{h-1}),
    \qquad h=1,\ldots,H.
  \end{equation*}
  Both chains start at $s_0$, so $\TV(\hat\mu_0,\mu_0)=0$, and unrolling the
  recursion gives
  \begin{equation*}
    \TV(\hat\mu_H,\mu_H)
    \le\varepsilon_{\mathrm{sim}}+\TV(\hat\mu_{H-1},\mu_{H-1})
    \le\cdots
    \le H\varepsilon_{\mathrm{sim}}+\TV(\hat\mu_0,\mu_0)
    =H\varepsilon_{\mathrm{sim}}.
  \end{equation*}
  Finally, $\TV\le1$ for any two probability measures.
\end{proof}

\begin{proof}[Proof of \Cref{thm:decomposition}]
  \emph{Value error.} Fix $(\ctx,a)$ and write $R(s)=R(\ctx,a,s)$ and
  $\hat R(s)=\hat R(\ctx,a,s)$, so that $\Qv(\ctx,a)=\E_{\mu_H}[R]$ and
  $\hat q(\ctx,a)=\E_{\hat\mu_H}[\hat R]$. Then
  \begin{align*}
    \abs{\hat q(\ctx,a)-\Qv(\ctx,a)}
    &=\Bigl|\E_{\hat\mu_H}\bigl[\hat R-R\bigr]+\bigl(\E_{\hat\mu_H}[R]-\E_{\mu_H}[R]\bigr)\Bigr|\\
    &\le\E_{\hat\mu_H}\abs{\hat R-R}+\abs{\E_{\hat\mu_H}[R]-\E_{\mu_H}[R]}
    &&\text{(triangle inequality)}\\
    &\le\varepsilon_{\mathrm{eval}}+\TV(\hat\mu_H,\mu_H)
    &&\text{(definition of $\varepsilon_{\mathrm{eval}}$; \cref{eq:tv-test} with $R\in[0,1]$)}\\
    &\le\varepsilon_{\mathrm{eval}}+\min\{1,H\varepsilon_{\mathrm{sim}}\}
    &&\text{(\Cref{lem:tv-composition})}.
  \end{align*}

  \emph{Selection.} Fix an instance with $\max_k\abs{q_k-\Qv_k}\le\varepsilon$,
  and let $k^\star\in\argmax_k\Qv_k$ and $\hat k\in\argmax_kq_k$, so that
  $q_{k^\star}\le q_{\hat k}$. Then
  \begin{align*}
    \Qv_{k^\star}-\Qv_{\hat k}
    &=\bigl(\Qv_{k^\star}-q_{k^\star}\bigr)+\bigl(q_{k^\star}-q_{\hat k}\bigr)+\bigl(q_{\hat k}-\Qv_{\hat k}\bigr)\\
    &\le\varepsilon+0+\varepsilon=2\varepsilon.
  \end{align*}
  Values lie in $[0,1]$, so the loss is also at most 1. If $\hat k$ does not
  maximize $\Qv$, then $\Qv_{\hat k}$ is at most the second-best value and the
  loss is at least $\Delta(X)$; hence $\Delta(X)>2\varepsilon$ forces
  $\Qv_{\hat k}=\Qv_{k^\star}$.

  \emph{Regret.} For a simulate-then-evaluate selector, the value error holds
  for every candidate with
  $\varepsilon=\min\{1,H\varepsilon_{\mathrm{sim}}\}+\varepsilon_{\mathrm{eval}}$,
  so
  \begin{equation*}
    \Reg=\E\bigl[\Qv_{k^\star}(X)-\Qv_{\hat k}(X)\bigr]
    \le2\bigl(\min\{1,H\varepsilon_{\mathrm{sim}}\}+\varepsilon_{\mathrm{eval}}\bigr).
  \end{equation*}
  For a direct value \wrm with $\sup_{X,k}\abs{q_k(X)-\Qv_k(X)}\le\varepsilon_w$,
  the same argument with $\varepsilon=\varepsilon_w$ gives $\Reg\le2\varepsilon_w$.
\end{proof}

\subsection{Objective mismatch in a restricted family}
\label{app:proof-mismatch}

\begin{proof}[Proof of \Cref{prop:mismatch}]
  \emph{True values.} Under candidate $a_k$ the outcome is
  $s'=(y,\xi_1,\ldots,\xi_d)$ with independent coordinates $y\sim\Bern(q_k)$
  and $\xi_j\sim\Bern(m_k)$, and the reward is $R(s')=y$. Hence
  $\Qv_k=\E[y]=q_k$, and $q_1>q_2$ makes $a_1$ the higher-value candidate.

  \emph{Likelihood fit.} The tied-rate family models all $1+d$ coordinates of
  candidate $k$ as independent $\Bern(\theta_k)$. The population
  log-likelihood of candidate $k$ is
  \begin{align*}
    \ell_k(\theta)
    &=\E\Bigl[y\log\theta+(1-y)\log(1-\theta)
      +\sum_{j=1}^d\bigl(\xi_j\log\theta+(1-\xi_j)\log(1-\theta)\bigr)\Bigr]\\
    &=(q_k+dm_k)\log\theta+(1+d-q_k-dm_k)\log(1-\theta).
  \end{align*}
  Its derivative
  \begin{equation*}
    \ell_k'(\theta)=\frac{q_k+dm_k}{\theta}-\frac{1+d-q_k-dm_k}{1-\theta}
  \end{equation*}
  is strictly decreasing on $(0,1)$ and vanishes at
  \begin{equation*}
    \hat\theta_k=\frac{q_k+dm_k}{1+d},
  \end{equation*}
  which is therefore the unique maximizer.

  \emph{Selection.} With the exact evaluator, the world model's value
  estimate is the model's mean of the reward bit,
  $\hat q_k=\E_{\hat\theta_k}[y]=\hat\theta_k$. Then
  \begin{equation*}
    \hat q_2-\hat q_1
    =\frac{(q_2+dm_2)-(q_1+dm_1)}{1+d}
    =\frac{d(m_2-m_1)-(q_1-q_2)}{1+d},
  \end{equation*}
  which is positive exactly when $d(m_2-m_1)>q_1-q_2$. In that case the
  selector chooses $a_2$ on every instance, with value regret
  $\Qv_1-\Qv_2=q_1-q_2$.

  \emph{Reward-only fit.} Fitting the same family to the reward bit alone
  maximizes $q_k\log\theta+(1-q_k)\log(1-\theta)$, whose unique maximizer is
  $\theta_k=q_k$ by the same derivative argument with $d=0$. The estimates are
  $(q_1,q_2)$, and the argmax is $a_1$.

  \emph{Example.} With $q=(0.8,0.3)$, $m=(0.1,0.9)$, and $d=10$,
  \begin{equation*}
    \hat\theta_1=\frac{0.8+10\cdot0.1}{11}=\frac{1.8}{11}\approx0.164,
    \qquad
    \hat\theta_2=\frac{0.3+10\cdot0.9}{11}=\frac{9.3}{11}\approx0.845,
  \end{equation*}
  and $d(m_2-m_1)=8>0.5=q_1-q_2$, so the selector picks $a_2$ with regret
  $0.8-0.3=0.5$.

  \emph{Winner probability.} For independent reward bits with ties broken
  uniformly at random,
  \begin{align*}
    \eta_1
    &=\Prob[y_1=1,\ y_2=0]+\tfrac12\,\Prob[y_1=y_2]\\
    &=q_1(1-q_2)+\tfrac12\bigl(q_1q_2+(1-q_1)(1-q_2)\bigr)\\
    &=\tfrac12+\tfrac12(q_1-q_2),
  \end{align*}
  which is $0.75$ in the example. Even the Bayes preference selector, which
  picks $a_1$, is right with probability $0.75$, and the likelihood-fitted
  world model, which picks $a_2$, is right with probability $0.25$. The
  construction concerns population objectives in this tied-rate family.
\end{proof}

\section{Idealized Gaussian Screening}
\label{app:screening}

Cheap judges let an agent screen more candidates. To ask how much breadth can
compensate for accuracy, we leave the bounded setting of \Cref{sec:setup} and
adopt the classical model of comparative judgment~\citep{thurstone1927law}.

\begin{assumption}[Idealized Gaussian screening]
  \label{ass:gaussian}
  On an instance, candidate values $V_k$ and fixed judge scores $S_k$ form
  independent and identically distributed pairs $(V_k,S_k)$, $k\in[K]$, each
  bivariate normal with value mean $\mu_V$, value spread $\sigma$, score mean
  $\mu_S$, score spread $\sigma_S$, and correlation $\rho\in[0,1]$.
\end{assumption}

\begin{lemma}[Pairwise accuracy determines correlation]
  \label{lem:sheppard}
  Under \Cref{ass:gaussian}, the pairwise accuracy of the judge is
  $p=\Prob[S_i>S_j\mid V_i>V_j]=\tfrac12+\tfrac1\pi\arcsin\rho$.
\end{lemma}

\begin{theorem}[Screening law]
  \label{thm:screening}
  Under \Cref{ass:gaussian}, selecting $\hat k=\argmax_kS_k$ yields
  \begin{equation}
    \E[V_{\hat k}]-\E[V_k]=\rho\,\sigma\,\eK,
    \qquad
    \eK=\E\bigl[\max\nolimits_{k\le K}Z_k\bigr],
    \label{eq:screening}
  \end{equation}
  where $Z_1,\ldots,Z_K$ are independent standard normal and $\eK$ is their
  expected maximum; $\eK$ increases in $K$, $e_2=1/\sqrt{\pi}$, and
  $\eK\le\sqrt{2\ln K}$.
\end{theorem}

\begin{corollary}[Breadth compensates slowly]
  \label{cor:breadth}
  A judge with correlation $\rho_W$ screening $K_W$ candidates matches the
  expected gain of one with $\rho_R>\rho_W$ screening $K_R$ only if
  $e_{K_W}\ge(\rho_R/\rho_W)e_{K_R}$, which requires
  $K_W\ge\exp\bigl((\rho_R/\rho_W)^2e_{K_R}^2/2\bigr)$.
\end{corollary}

\begin{figure}[ht]
  \centering
  \includegraphics[width=0.56\linewidth]{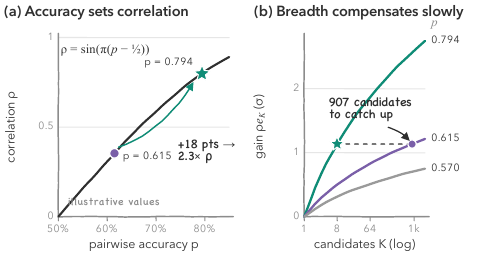}
  \caption{\textbf{In the idealized Gaussian model, breadth compensates for
    accuracy only slowly.}
    (a)~\Cref{lem:sheppard} maps pairwise accuracy $p$ to correlation $\rho$.
    (b)~Expected gain $\rho\,\eK$ in units of $\sigma$
    (\Cref{thm:screening}) for illustrative $p=0.615$ and $p=0.794$: the
    $p=0.615$ judge needs $K=907$ candidates to match the $p=0.794$ judge at
    $K=8$. Curves are analytic and are not fitted to the agents studied
    here.}
  \label{fig:screening}
\end{figure}

\begin{figure}[ht]
  \centering
  \includegraphics[width=0.66\linewidth]{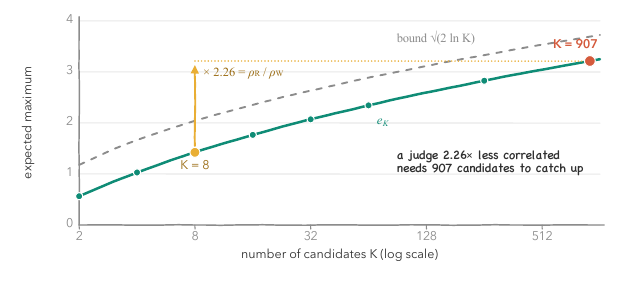}
  \caption{\textbf{The expected maximum of $K$ standard normals grows like
    $\sqrt{\ln K}$.} $\eK=\int zK\phi(z)\Phi(z)^{K-1}\,dz$ by numerical
    integration, with $\phi$ and $\Phi$ the standard normal density and
    distribution function, against the bound $\sqrt{2\ln K}$ of
    \Cref{thm:screening}. Multiplying $K$ by 128, from 8 to 1024, multiplies
    $\eK$ by 2.3. The marked points give the catch-up example: $e_{907}$ is
    $\rho_R/\rho_W=2.26$ times $e_8$.}
  \label{fig:ek}
\end{figure}

\Cref{thm:screening} is the breeder's equation of quantitative
genetics~\citep{falconer1996introduction}: response to selection equals
intensity times accuracy times spread. For illustrative pairwise accuracies of
0.615 and 0.794, the levels reported for an LLM judge and measured for \fcm on
different test sets, \Cref{lem:sheppard} gives $\rho=0.35$ and $0.80$, and the
first judge needs at least 175 and exactly 907 candidates to match what the
second obtains from 8 (\Cref{fig:screening,fig:ek}). The conclusion depends on
the idealization: Gaussian values are unbounded, a pairwise selector need not
define a global score, accuracy against a noisy winner is not accuracy against
expected value, and whether breadth pays also depends on the cost of
generating candidates.

\subsection{Proofs}
\label{app:proof-screening}

\begin{proof}[Proof of \Cref{lem:sheppard}]
  Fix $i\ne j$ and let $U=V_i-V_j$ and $W=S_i-S_j$. The pairs $(V_i,S_i)$ and
  $(V_j,S_j)$ are independent and identically distributed, so
  \begin{align*}
    \Var(U)&=\Var(V_i)+\Var(V_j)=2\sigma^2,\qquad
    \Var(W)=\Var(S_i)+\Var(S_j)=2\sigma_S^2,\\
    \Cov(U,W)&=\Cov(V_i,S_i)+\Cov(V_j,S_j)-\Cov(V_i,S_j)-\Cov(V_j,S_i)
              =2\rho\sigma\sigma_S-0-0.
  \end{align*}
  Hence $(U,W)$ is a centered bivariate normal vector with correlation
  \begin{equation*}
    \frac{\Cov(U,W)}{\sqrt{\Var(U)\Var(W)}}
    =\frac{2\rho\sigma\sigma_S}{\sqrt{2\sigma^2\cdot2\sigma_S^2}}=\rho.
  \end{equation*}
  Sheppard's formula~\citep{sheppard1899application} for such a vector gives
  $\Prob[U>0,W>0]=\tfrac14+\tfrac1{2\pi}\arcsin\rho$, and $\Prob[U>0]=\tfrac12$
  by symmetry. Therefore
  \begin{equation*}
    p=\Prob[W>0\mid U>0]
    =\frac{\Prob[U>0,\ W>0]}{\Prob[U>0]}
    =\frac{\tfrac14+\tfrac1{2\pi}\arcsin\rho}{\tfrac12}
    =\frac12+\frac1\pi\arcsin\rho.
    \qedhere
  \end{equation*}
\end{proof}

\begin{proof}[Proof of \Cref{thm:screening}]
  \emph{Conditional mean.} Set $Z_k=(S_k-\mu_S)/\sigma_S$, so that
  $Z_1,\ldots,Z_K$ are independent standard normal. For a bivariate normal
  pair,
  \begin{equation*}
    \E[V_k\mid S_k]=\mu_V+\rho\,\frac{\sigma}{\sigma_S}\,(S_k-\mu_S)
    =\mu_V+\rho\sigma Z_k.
  \end{equation*}
  The pairs are independent across candidates, so $V_k$ is independent of
  $(S_j)_{j\ne k}$ given $S_k$, and
  $\E[V_k\mid S_1,\ldots,S_K]=\E[V_k\mid S_k]$.

  \emph{Selected value.} The index $\hat k=\argmax_kS_k=\argmax_kZ_k$ is a
  function of the scores, so
  \begin{align*}
    \E\bigl[V_{\hat k}\mid S_1,\ldots,S_K\bigr]
    &=\sum_{k=1}^K\Ind\{\hat k=k\}\,\E\bigl[V_k\mid S_1,\ldots,S_K\bigr]\\
    &=\sum_{k=1}^K\Ind\{\hat k=k\}\,\bigl(\mu_V+\rho\sigma Z_k\bigr)\\
    &=\mu_V+\rho\sigma Z_{\hat k}
     =\mu_V+\rho\sigma\max_{k\le K}Z_k.
  \end{align*}
  Taking expectations gives $\E[V_{\hat k}]=\mu_V+\rho\sigma\eK$, and
  $\E[V_k]=\mu_V$ for every fixed $k$, which proves \cref{eq:screening}.

  \emph{Monotonicity.} $\max_{k\le K+1}Z_k\ge\max_{k\le K}Z_k$ pointwise, so
  $e_{K+1}\ge\eK$.

  \emph{Two candidates.} Using $\max(a,b)=\tfrac12(a+b)+\tfrac12\abs{a-b}$ and
  $Z_1-Z_2\sim\cN(0,2)$, whose absolute value has mean $\sqrt2\cdot\sqrt{2/\pi}$,
  \begin{equation*}
    e_2=\tfrac12\E[Z_1+Z_2]+\tfrac12\E\abs{Z_1-Z_2}
    =0+\tfrac12\cdot\frac{2}{\sqrt\pi}=\frac1{\sqrt\pi}.
  \end{equation*}

  \emph{Upper bound.} For every $t>0$,
  \begin{align*}
    \exp(t\,\eK)
    &\le\E\Bigl[\exp\bigl(t\max_kZ_k\bigr)\Bigr]
    &&\text{(Jensen's inequality, $\exp$ convex)}\\
    &=\E\Bigl[\max_k\exp(tZ_k)\Bigr]
    \le\sum_{k=1}^K\E\bigl[\exp(tZ_k)\bigr]
    &&\text{(a maximum of nonnegative terms is at most their sum)}\\
    &=K\exp(t^2/2)
    &&\text{(moment generating function of $\cN(0,1)$)}.
  \end{align*}
  Taking logarithms and dividing by $t$, $\eK\le(\ln K)/t+t/2$. The right-hand
  side is minimized at $t=\sqrt{2\ln K}$, where it equals $\sqrt{2\ln K}$.
\end{proof}

\begin{proof}[Proof of \Cref{cor:breadth}]
  By \Cref{thm:screening}, the two judges gain $\rho_W\sigma e_{K_W}$ and
  $\rho_R\sigma e_{K_R}$ over the average candidate. The first matches the
  second exactly when
  \begin{equation*}
    \rho_W\sigma e_{K_W}\ge\rho_R\sigma e_{K_R}
    \iff e_{K_W}\ge\frac{\rho_R}{\rho_W}\,e_{K_R}.
  \end{equation*}
  Since $e_{K_W}\le\sqrt{2\ln K_W}$, this requires
  $\sqrt{2\ln K_W}\ge(\rho_R/\rho_W)\,e_{K_R}$, that is,
  $\ln K_W\ge(\rho_R/\rho_W)^2e_{K_R}^2/2$; exponentiating gives the bound.
\end{proof}

\begin{lemma}[Knockout compounding]
  \label{lem:bracket}
  In a single-elimination bracket over $K=2^m$ candidates, suppose that, in
  every round and conditionally on having survived, the best candidate wins
  its match with probability $p$, independently of earlier rounds. Then it is
  selected with probability $p^m=K^{\log_2p}$.
\end{lemma}

\begin{proof}
  Let $B_r$ be the event that the best candidate wins its match in round $r$.
  It is selected exactly when $B_1\cap\cdots\cap B_m$ occurs, so by the chain
  rule and the assumption,
  \begin{equation*}
    \Prob\bigl[B_1\cap\cdots\cap B_m\bigr]
    =\prod_{r=1}^m\Prob\bigl[B_r\bigm| B_1\cap\cdots\cap B_{r-1}\bigr]
    =p^m=2^{m\log_2p}=K^{\log_2p}.
    \qedhere
  \end{equation*}
\end{proof}

The conditional accuracy in \Cref{lem:bracket} is the accuracy of the best
candidate against its actual opponents, which generally differs from the
average pairwise accuracy over all pairs; the lemma is an idealization, not a
prediction for the tournaments of \Cref{sec:method-agent}.

\subsection{Numerical values}
\label{app:numerics}

\paragraph{Catch-up example.}
For illustrative pairwise accuracies $p_W=0.615$ and $p_R=0.794$,
\Cref{lem:sheppard} gives $\rho_W=0.3535$ and $\rho_R=0.7978$. Matching the
$p_R$ judge at $K_R=8$ requires $e_{K_W}\ge(\rho_R/\rho_W)\,e_8=3.2131$.
\Cref{cor:breadth} gives $K_W\ge174.5$, so at least 175; numerically
$e_{906}=3.21298$ and $e_{907}=3.21330$, so the smallest sufficient $K_W$ is
907. A Monte Carlo check with 100,000 draws per setting, for
$\rho\in\{0,0.35,0.8\}$ and $K\in\{2,8,32\}$, matches \cref{eq:screening} and
\Cref{lem:sheppard} within five standard errors; this verifies the formulas,
not their fit to real agents.

\section{Data, Training, and Inference Details}
\label{app:details}

\paragraph{Partitions.}
We split the 18,438 source pairs of the corpus once, before training, into
five partitions: 5,000 pairs for training, 250 for development, 250 for
calibration, 500 for an internal test, and 12,438 for the final test. No
source pair, and no identical unordered pair of programs, appears in two
partitions; tasks are shared across partitions, and a program can appear in
pairs of different partitions. Development pairs select the checkpoint,
calibration pairs fit the temperature, and the internal test is a check run
once; the final test carries the main results, scored once with the full
input and once without the data report. Of its 12,438 pairs, 5,996 contain at
least one program absent from training and 501 contain two. A few identical
program pairs carry conflicting labels in the source corpus; we keep them as
they are.

\paragraph{Model and optimization.}
\fcm keeps the backbone of the released Kev-9B, Qwen3.5-9B-Base, frozen in
BF16, and initializes the LoRA adapter (rank 16, $\alpha=32$, dropout 0.05)
and the pointer head from Kev-9B~\citep{palmer2026kev}; only these two parts
are trained. The pointer head consists of the two affine projections of
\cref{eq:pointer}, from 4,096 to 256 dimensions. Each training pair is
rendered in three input formats (full JSON, JSON without the data report, and
sectioned text) and in both candidate orders, and each epoch draws one of the
six variants per pair, so the model sees every pair once per epoch in a
varying surface form. We minimize \cref{eq:loss} with AdamW (weight decay
0.01, gradient clipping at 1.0) at learning rates of $10^{-5}$ for the adapter
and $2\times10^{-5}$ for the head, with 5\% linear warm-up and cosine decay to
10\% of the peak. An effective batch of 32 pairs (micro-batch 1, gradient
accumulation 4, eight GPUs) gives 628 updates over four epochs. Inputs are
never truncated; the longest has about 46K tokens. Training on eight A800
GPUs takes 4.18 hours, including loading and validation.

\paragraph{Selection and calibration.}
We keep the checkpoint with the highest development accuracy, breaking ties
by lower NLL; this is the final, fourth-epoch checkpoint, at 78.4\%
development accuracy. We then fit the temperature $T$ of \cref{eq:pointer} by
minimizing NLL on the calibration partition over a logarithmic grid, which
gives $T=1.086$ for \fcm and $T=12.29$ for Kev-9B. Neither step uses
final-test labels.

\paragraph{Input format.}
A request places all content in a state and asks one question with two short
options. The following synthetic request illustrates the schema; real
candidates are complete programs.
\begin{small}
\begin{verbatim}
state: {"task": "synthetic-binary-classification",
        "task_description": "Predict a binary target. Higher ROC-AUC is better.",
        "data_analysis_report": "Numeric features with missing values.",
        "solutions": {"solution_0": "<complete program A>",
                      "solution_1": "<complete program B>"}}
question: Which solution is expected to perform better?
options:  solution_0: Solution 0 is expected to outperform Solution 1.
          solution_1: Solution 1 is expected to outperform Solution 0.
\end{verbatim}
\end{small}
The response is the chosen option with a probability for each option. Main
results score the original order once; swapping is used only as training
augmentation and in the order analysis of \Cref{app:transfer}.

\section{Additional Results}
\label{app:additional}

\subsection{Shared-pair comparison with Qwen and an earlier adaptation}

\begin{figure}[ht]
  \centering
  \includegraphics[width=0.75\linewidth]{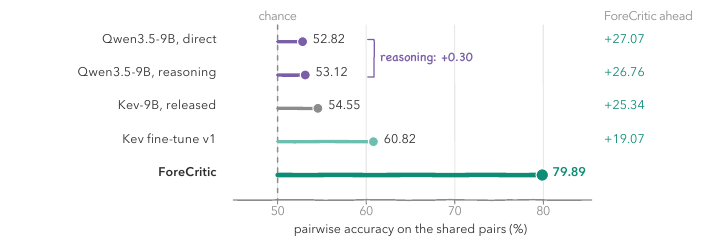}
  \caption{\textbf{On the 9,879 pairs common to all evaluations, only the
    execution-adapted models clearly exceed chance.} Pairwise accuracy on the
    same pairs and labels; the models differ in weights, inputs, and
    training. Kev fine-tune v1 is an earlier adaptation with a different
    recipe and split. Labels on the right give \fcm's paired advantage, which
    is significant against every model.}
  \label{fig:shared}
\end{figure}

\Cref{fig:shared} places the two Qwen judges, the released Kev-9B, and an
earlier adaptation on the 9,879 final-test pairs that every evaluation
covers. Reasoning raises Qwen3.5-9B from 52.82\% to 53.12\%, and neither
adaptation trained on these pairs. The earlier adaptation reaches 60.82\%;
the recipe of \Cref{sec:method-training} reaches 79.89\%.

\subsection{Probability quality}

\begin{figure}[ht]
  \centering
  \includegraphics[width=\linewidth]{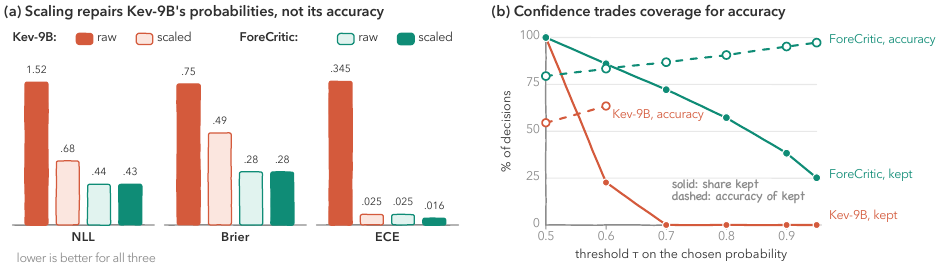}
  \caption{\textbf{Temperature scaling refines an already calibrated \fcm and
    repairs Kev-9B's probabilities but not its accuracy; \fcm's confidence
    trades coverage for accuracy.} Final test, 12,438 pairs. (a)~NLL, Brier
    score summed over the two classes, and ECE over ten bins, before and after
    temperature scaling; accuracy is unchanged by scaling (54.53\% for
    Kev-9B, 79.44\% for \fcm). (b)~Fraction of decisions kept and their
    accuracy when only decisions with chosen-class probability at least
    $\tau$ are accepted; thresholds were not tuned on these labels.}
  \label{fig:probquality}
\end{figure}

Temperature scaling lowers Kev-9B's NLL from 1.525 to 0.685 and its ECE from
0.345 to 0.025, but its accuracy stays at 54.53\%, and above $\tau=0.6$ it
accepts no decision at all (\Cref{fig:probquality}). For \fcm, scaling changes
NLL only from 0.4364 to 0.4332 and ECE from 0.0251 to 0.0163, because the raw
probabilities are already close to calibrated. Raising $\tau$ from 0.5 to 0.95
shrinks \fcm's accepted share from 100\% to 25.2\% and raises its accuracy
from 79.4\% to 97.3\%.

\subsection{Transfer details}
\label{app:transfer}

The transfer candidates are the 48 programs with a public validation metric
that \foreagent executed with \fcm as its selector on ten MLE-Dojo tasks; one
task, \texttt{random-acts-of-pizza}, also occurs in \fcm's training corpus.
Both judges receive the same fields and question; \fcm reads compact JSON, and
DeepSeek-V4.1-Flash reads the fields as text with temperature 0, at most 200
output tokens, and reasoning disabled. The accuracy difference stays
insignificant when whole tasks are resampled. Probabilities recovered from the
logged confidences have an approximate ECE of 0.037 without refitting the
temperature, and a threshold of 0.7 keeps 90 of 182 judgments (49.5\%) with 72
correct (80.0\%).

\begin{figure}[ht]
  \centering
  \includegraphics[width=0.9\linewidth]{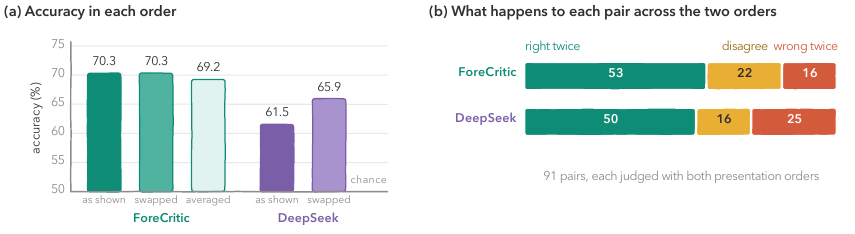}
  \caption{\textbf{On fresh agent candidates, \fcm is more accurate in each
    presentation order but less order-consistent than DeepSeek.} 91 pairs
    from ten MLE-Dojo tasks. (a)~Accuracy in each order and, for \fcm, after
    averaging the two orders' probabilities. (b)~Each pair judged in both
    orders is right twice, right once (the two orders disagree), or wrong
    twice.}
  \label{fig:transfer-orders}
\end{figure}

\paragraph{Order consistency and averaging.}
After mapping both orders back to the same programs, \fcm makes the same
choice on 69 of 91 pairs (75.8\%): 53 are right in both orders, 16 wrong in
both, and 22 split (\Cref{fig:transfer-orders}); DeepSeek is consistent on
82.4\%. If the presentation order carries no information about the winner,
the swap $\sigma$ of the two candidates satisfies $\eta(\sigma X)=\sigma\eta(X)$.
The order-averaged predictor $\bar w(X)=\tfrac12[w(X)+\sigma w(\sigma X)]$ then
satisfies
\begin{align*}
  \E\bigl[-\log\bar w_Y(X)\bigr]
  &=\E\Bigl[-\log\Bigl(\tfrac12w_Y(X)+\tfrac12\bigl(\sigma w(\sigma X)\bigr)_Y\Bigr)\Bigr]\\
  &\le\tfrac12\E\bigl[-\log w_Y(X)\bigr]
   +\tfrac12\E\bigl[-\log\bigl(\sigma w(\sigma X)\bigr)_Y\bigr]
  &&\text{(convexity of $-\log$)},
\end{align*}
so its excess cross-entropy, and with it the bound of \Cref{prop:proper}, is
no worse than the mean over the two orders. Averaging \fcm's two probabilities
per pair yields 63 of 91 correct (69.23\%), an insignificant change of $-1.10$
points from the mean single-order accuracy; in this sample, order averaging
neither helps nor clearly hurts accuracy, at the price of a second forward
pass.

\section{Agent Integration Details}
\label{app:agents}

\paragraph{Settings.}
All runs use DeepSeek-V4.1-Flash for code generation with temperature 0.5, at
most 20 generations per step, a 6-hour task limit, a 3,600-second execution
timeout, and a final repair budget of 2. AIDE and ML-Master start from 5
drafts and search with a verification budget of 0; \foreagent starts from 3
drafts with batches of 10 and threshold $\tau=0.7$ on the chosen-class
probability, judging each pair in one order with the first position
alternating across comparisons (\Cref{alg:tournament}). AIDE and \foreagent
use a fixed seed; ML-Master does not expose one. Each configuration
contributes one scored run per task.

\begin{algorithm}[ht]
  \caption{Confidence-gated pairwise tournament with a \wrm (\foreagent integration)}
  \label{alg:tournament}
  \begin{algorithmic}[1]
    \REQUIRE context $X$, batch $B$ of proposed programs, \wrm $w$, threshold $\tau$
    \WHILE{$\abs{B}>1$}
      \STATE pair the programs in $B$;\quad $E\leftarrow\emptyset$
      \FOR{each pair $(a,b)$}
        \STATE $(p_a,p_b)\leftarrow w(X,(a,b))$ \COMMENT{one forward pass}
        \STATE add $p_a,p_b$ to the soft scores of $a$ and $b$
        \IF{$\max(p_a,p_b)\ge\tau$}
          \STATE add the program with the smaller probability to $E$
        \ENDIF
      \ENDFOR
      \IF{$E=\emptyset$} \STATE \textbf{break} \COMMENT{no confident elimination} \ENDIF
      \STATE $B\leftarrow B\setminus E$
    \ENDWHILE
    \STATE execute the program of $B$ with the highest soft score
  \end{algorithmic}
\end{algorithm}

\paragraph{Cost accounting.}
Task time is a run's elapsed wall-clock time from start to scored submission.
It includes the data report, code generation, every selector call, and the
final execution with up to two repairs, and it excludes loading \fcm, which
the service does once before the runs. The 446 \fcm comparisons in the 30 runs
took about 560 seconds of logged service time in total, including queueing
(median 1.2 seconds). Each comparison is one prefill over the task, report,
and both programs, with no decoding. One-time costs lie outside task time:
adaptation took 4.18 hours on eight A800 GPUs (33.4 GPU-hours), and the 5,000
training labels reuse executions from the Predict-Before-Execute corpus. Code
generation calls the same model in every configuration.

\begin{figure}[ht]
  \centering
  \includegraphics[width=\linewidth]{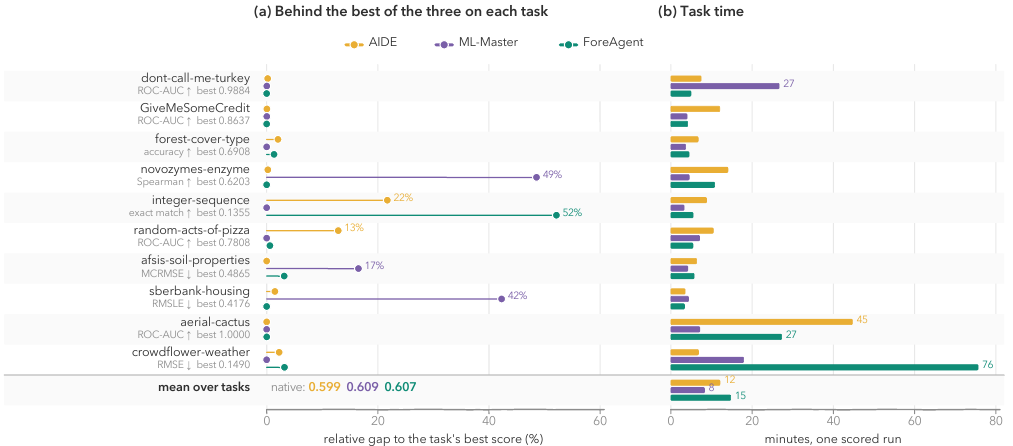}
  \caption{\textbf{With \fcm as selector, the three agents are within 4\% of
    each other on five of ten tasks, and a few long tasks dominate task time.}
    One scored run per task. (a)~Relative gap between each agent's native
    score and the best of the three on that task, taking the metric direction
    into account (0 is the best); under each task are its metric, direction,
    and best score, and the bottom row gives the mean native scores of
    \Cref{tab:agents}. (b)~Task time in minutes.}
  \label{fig:pertask}
\end{figure}

\paragraph{Per-task results and re-scoring.}
We re-scored all 30 \fcm submissions independently against the private test
answers (\Cref{fig:pertask}). Averaged over tasks, the native scores are
0.5994 for AIDE, 0.6088 for ML-Master, and 0.6068 for \foreagent. A few long
tasks account for much of the time: \texttt{aerial-cactus} for AIDE (45
minutes) and \foreagent (27 minutes), and \texttt{crowdflower-weather} for
\foreagent (76 minutes).

\paragraph{Original and Jev selectors.}
The original-selector and Jev rows of \Cref{tab:agents} use the same agent
configuration, tasks, and scoring as the \fcm runs. In the original AIDE and
ML-Master, candidates are executed during search; in the original \foreagent,
an LLM judge settles the tournament. We report their mean native scores and
mean task times, which follow the same definitions; their per-task values are
not reproduced here. Because native metrics differ in scale and direction,
these means summarize each configuration rather than measure a common
quantity, and with one run per task we do not test the differences for
significance.